\documentclass[conference]{IEEEtran}
\IEEEoverridecommandlockouts
\usepackage{cite}
\usepackage{amsmath,amssymb,amsfonts}
\usepackage{algorithmic}
\usepackage{graphicx}
\usepackage{textcomp}
\usepackage{hyperref}
\usepackage{xcolor}
\usepackage{subcaption}
\usepackage{multirow}

\def\BibTeX{{\rm B\kern-.05em{\sc i\kern-.025em b}\kern-.08em
    T\kern-.1667em\lower.7ex\hbox{E}\kern-.125emX}}
\begin{document}


\newcommand{\bestTestScore}{1.298}

\newcommand{\samil}[1]{\textcolor{red}{#1}}

\title{ShelfRectNet: Single View Shelf Image Rectification with Homography Estimation\\}


\author{\IEEEauthorblockN{1\textsuperscript{st} Onur Berk Tore}
\IEEEauthorblockA{\textit{Research \& Development} \\
\textit{REM People}\\
Istanbul, Türkiye \\
onurberk.tore@rempeople.com}
\and
\IEEEauthorblockN{2\textsuperscript{nd}Ibrahim Samil Yalciner}
\IEEEauthorblockA{\textit{Research \& Development} \\
\textit{REM People}\\
Istanbul, Türkiye \\
samil.yalciner@rempeople.com}
\and
\IEEEauthorblockN{3\textsuperscript{rd} Server Calap}
\IEEEauthorblockA{\textit{Research \& Development} \\
\textit{REM People}\\
Istanbul, Türkiye \\
server.calap@rempeople.com}
}

\maketitle
\IEEEpeerreviewmaketitle

\begin{abstract}
Estimating homography from a single image remains a challenging yet practically valuable task, particularly in domains like retail, where only one viewpoint is typically available for shelf monitoring and product alignment. In this paper, we present a deep learning framework that predicts a 4-point parameterized homography matrix to rectify shelf images captured from arbitrary angles. Our model leverages a ConvNeXt-based backbone for enhanced feature representation and adopts normalized coordinate regression for improved stability. To address data scarcity and promote generalization, we introduce a novel augmentation strategy by modeling and sampling synthetic homographies. Our method achieves a mean corner error of \bestTestScore{}  pixels on the test set. When compared with both classical computer vision and deep learning-based approaches, our method demonstrates competitive performance in both accuracy and inference speed. Together, these results establish our approach as a robust and efficient solution for real-world single-view rectification. To encourage further research in this domain, we will make our dataset, ShelfRectSet, and code publicly available.
\end{abstract}

\begin{IEEEkeywords}
shelf, rectification, warping, fronto-parallel, homography
\end{IEEEkeywords}

\section{Introduction}
\begin{figure}[t]
    \centering
    \begin{subfigure}[b]{0.20\textwidth}
        \centering
        \includegraphics[width=\textwidth]{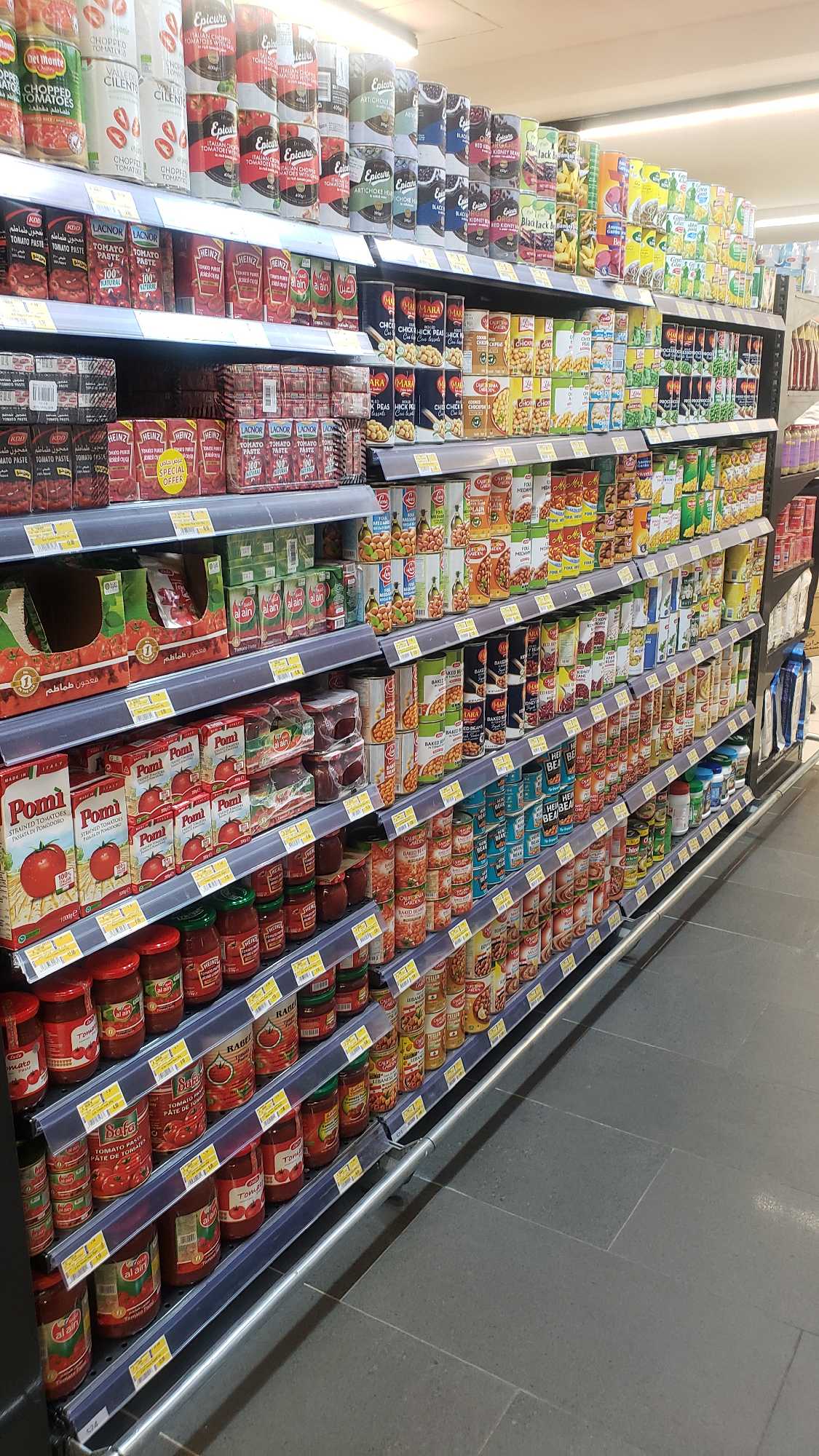}
    \end{subfigure}
    \hfill
    \begin{subfigure}[b]{0.2665\textwidth}
        \centering
        \includegraphics[width=\textwidth]{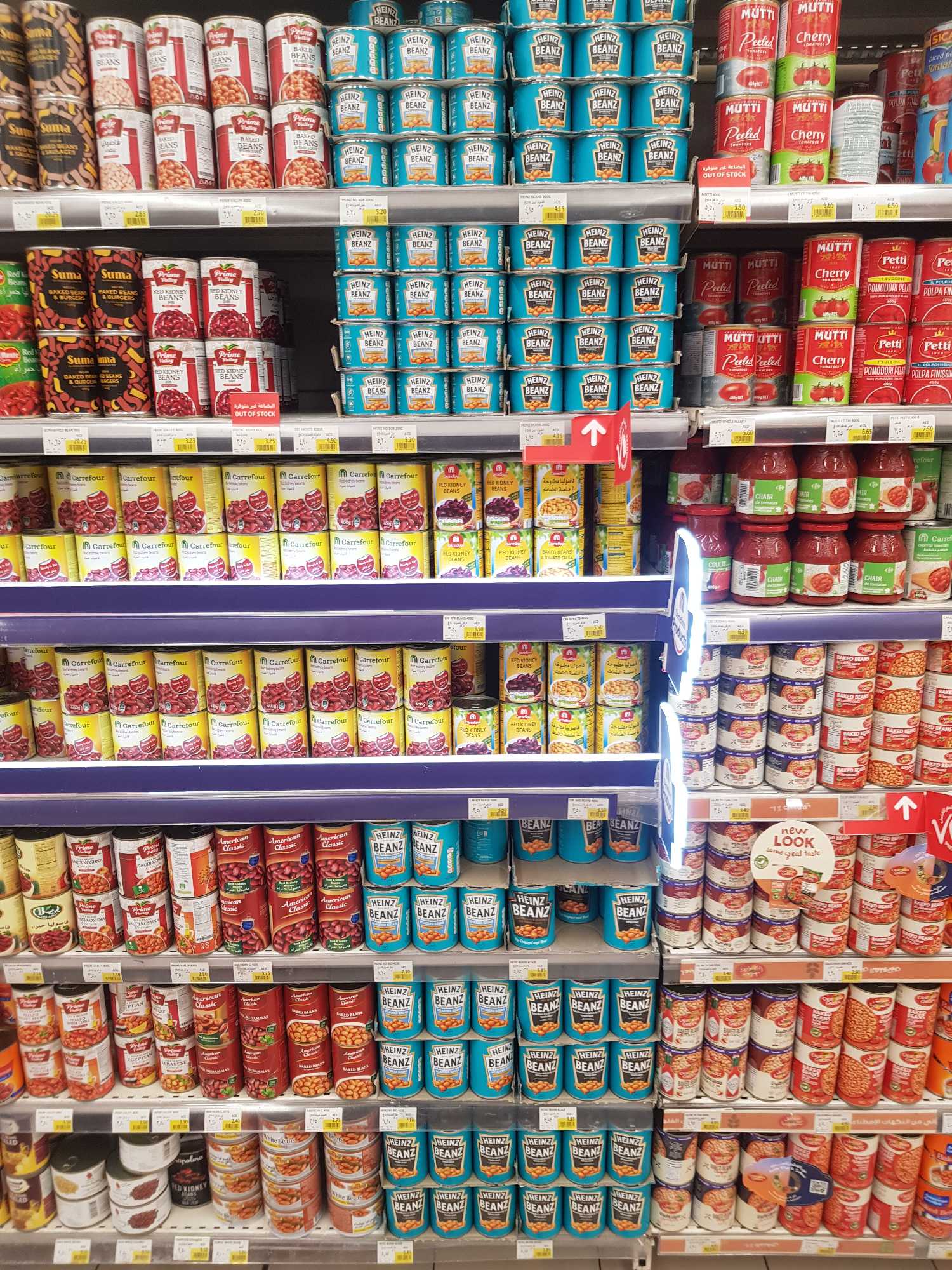}
    \end{subfigure}
    \caption{Two representative retail shelf images, one exhibiting significant perspective distortion (left) and the other captured from a fronto-parallel viewpoint (right).}

    \label{fig:shelf_distortion_example}
\end{figure}

Homography estimation from a single image is a challenging yet practically valuable task in computer vision. A critical application is the rectification of retail shelf images, where store employees or auditors often capture photos from a single, convenient, but arbitrary viewpoint. This results in significant perspective distortion, as illustrated in Fig. \ref{fig:shelf_distortion_example}, which complicates automated analysis. The core challenge is to computationally correct this distortion by estimating a homography that transforms the skewed image into a normalized, fronto-parallel view, making products appear as if photographed head-on.

This rectification step is vital for the retail domain, as it standardizes the visual data for subsequent machine learning pipelines. By creating a consistent, straight-on view of products, it dramatically improves the accuracy and reliability of automated tasks such as out-of-stock detection, planogram compliance checks, product recognition, and inventory management. Addressing the real-world constraint of working with a single, unpaired image is the primary motivation for our work.

The underlying task of homography estimation has traditionally been approached from two perspectives. Classical methods rely on detecting and matching sparse features between two images, but these techniques often fail in scenes with low texture or significant changes in perspective and lighting. To overcome these limitations, modern approaches have turned to deep learning, using complex Convolutional Neural Network (CNN) \cite{lecun1998gradient} and transformer \cite{dosovitskiy2020image} architectures to predict the homography between two images. The evolution of deep, two-view homography estimation began with the pioneering end-to-end model DeepImageHomographyNet \cite{detone2016deep}, which uses a CNN to directly regress the eight numerical values of a 4-point parameterization from a pair of stacked images.

Subsequent research has focused on improving this foundation in several key areas. To create more efficient and lightweight models, some have adopted ShuffleNet-style architectures that incorporate photometric loss \cite{wang2019efficient} or have used ShuffleNet V2 compression units to reduce model size \cite{chen2021fast}. Others have sought higher accuracy through hybrid frameworks that combine a CNN's initial prediction with a subsequent refinement step based on photometric error minimization \cite{Kang2019Combining}.

More recently, the field has moved towards even more sophisticated architectures. One major trend is the development of fully trainable iterative and recurrent networks that progressively refine the homography estimate \cite{cao2022iterative, Cao2023Recurrent}. Simultaneously, transformers have been integrated to better capture feature correlations, whether for cross-resolution challenges \cite{shao2021localtrans}, large-baseline scenes \cite{Zhou2023Deep}, or general correspondence \cite{Li2023Multi}. Specialized methods have also been developed to handle difficult scenarios like dynamic content in videos \cite{Le2020Deep, Yang2020} or to improve the realism of training data \cite{Jiang2023Supervised}.

However, a common thread among these advanced methods is their reliance on paired views, making them unsuitable for our target application. While automatic single-image rectification methods were developed even before the rise of deep learning such as \cite{chaudhury2013autorectification}, restores the parallelism of lines by first estimating two vanishing points directly from edgelets and then computing a rectification homography, there remains a need for a robust, data-driven approach.

To address this gap with a modern approach, we propose a novel deep learning framework for single-image homography estimation tailored to retail shelf rectification.  Our approach uses a ConvNeXt-based backbone \cite{liu2022convnet} for feature extraction, a modernized pure ConvNet that incorporates design elements from Vision Transformers to achieve superior performance while maintaining high efficiency. Our network predicts a 4-point parameterization \cite{detone2016deep} of the homography matrix. Unlike prior work that relies heavily on large, hand-curated datasets or multi-view supervision, we employ a targeted augmentation strategy that samples realistic homography perturbations based on the observed distribution of corner displacements in our data. This method increases the range of viewpoint variations encountered during training without requiring additional manual annotation, empirically reducing overfitting by exposing the model to a broader spectrum of geometric distortions. Our experimental evaluation demonstrates the effectiveness of the proposed framework, as we achieve a mean corner error of \bestTestScore{} pixels on a held-out test set of real-world shelf images, outperforming traditional methods.

In summary, our key contributions are as follows:

\begin{itemize}

\item We present a deep learning model for estimating rectifying homographies from single-view images, with a focus on retail shelf scenarios.

\item We introduce a novel data augmentation strategy that samples homographies from the training distribution to simulate diverse perspective distortions.

\item We introduce a new, large-scale dataset, ShelfRectSet, of annotated retail shelf images and will make it publicly available to encourage further research.

\end{itemize}

By addressing the challenge of single-image rectification in a structured, real-world setting, this work expands the scope of homography estimation beyond traditional assumptions and contributes to the growing intersection of geometric learning and domain-specific computer vision.

\section{Dataset Generation and Augmentation 
Pipeline}

Our methodology is developed using a dataset comprised of images captured in real-world retail environments. These images are acquired from a single, unconstrained viewpoint, and as a result, they inherently exhibit significant perspective distortion.

\begin{figure}[htbp]
\centerline{\includegraphics[width=0.5\textwidth]{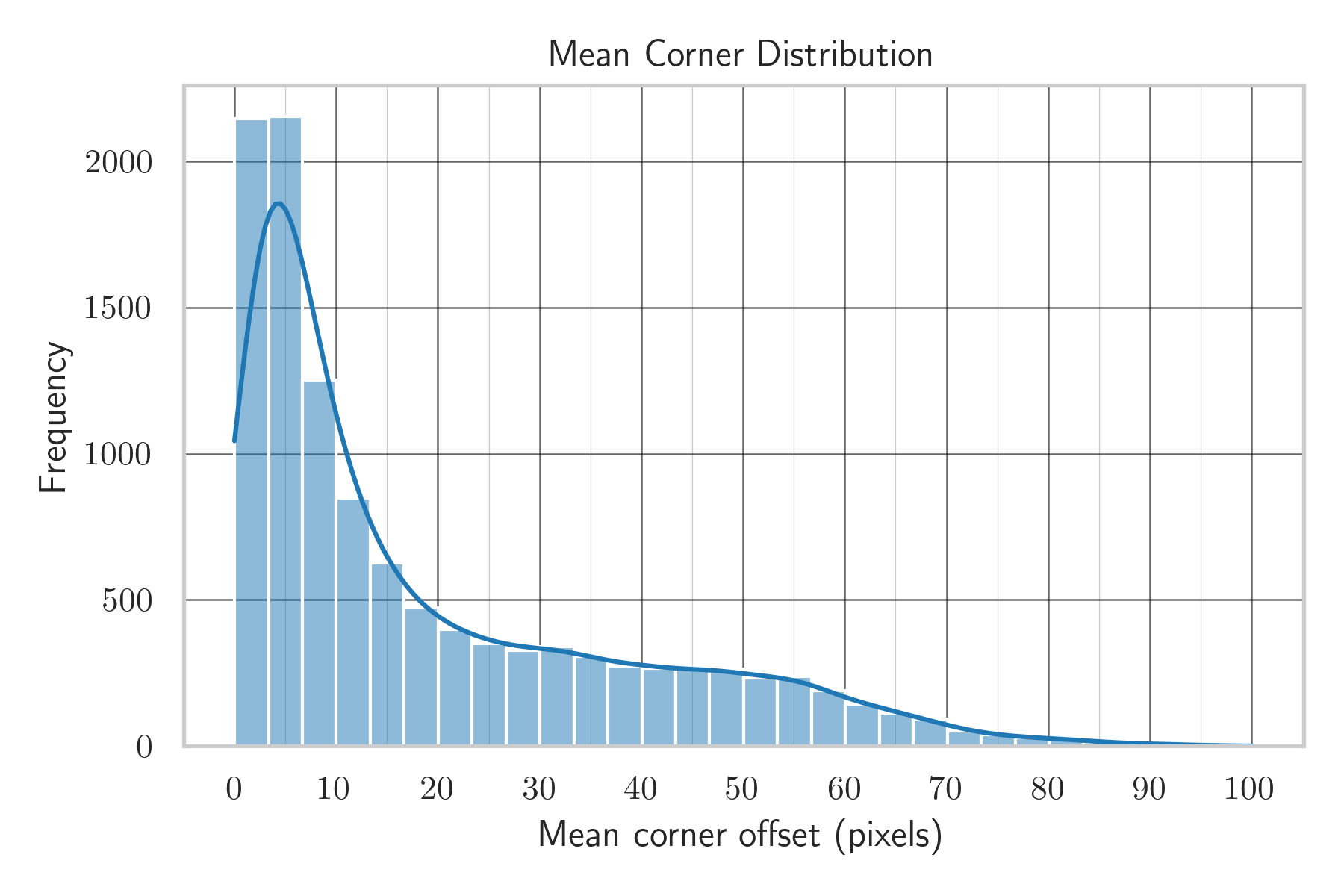}}
   \caption{Mean displacement values (in pixels) for each of the four corners across the entire ShelfRectSet, illustrating the distribution of annotation adjustments.}
   \label{fig:mean_displacement}
\end{figure}

\begin{figure}[htbp]
\centerline{\includegraphics[width=0.5\textwidth]{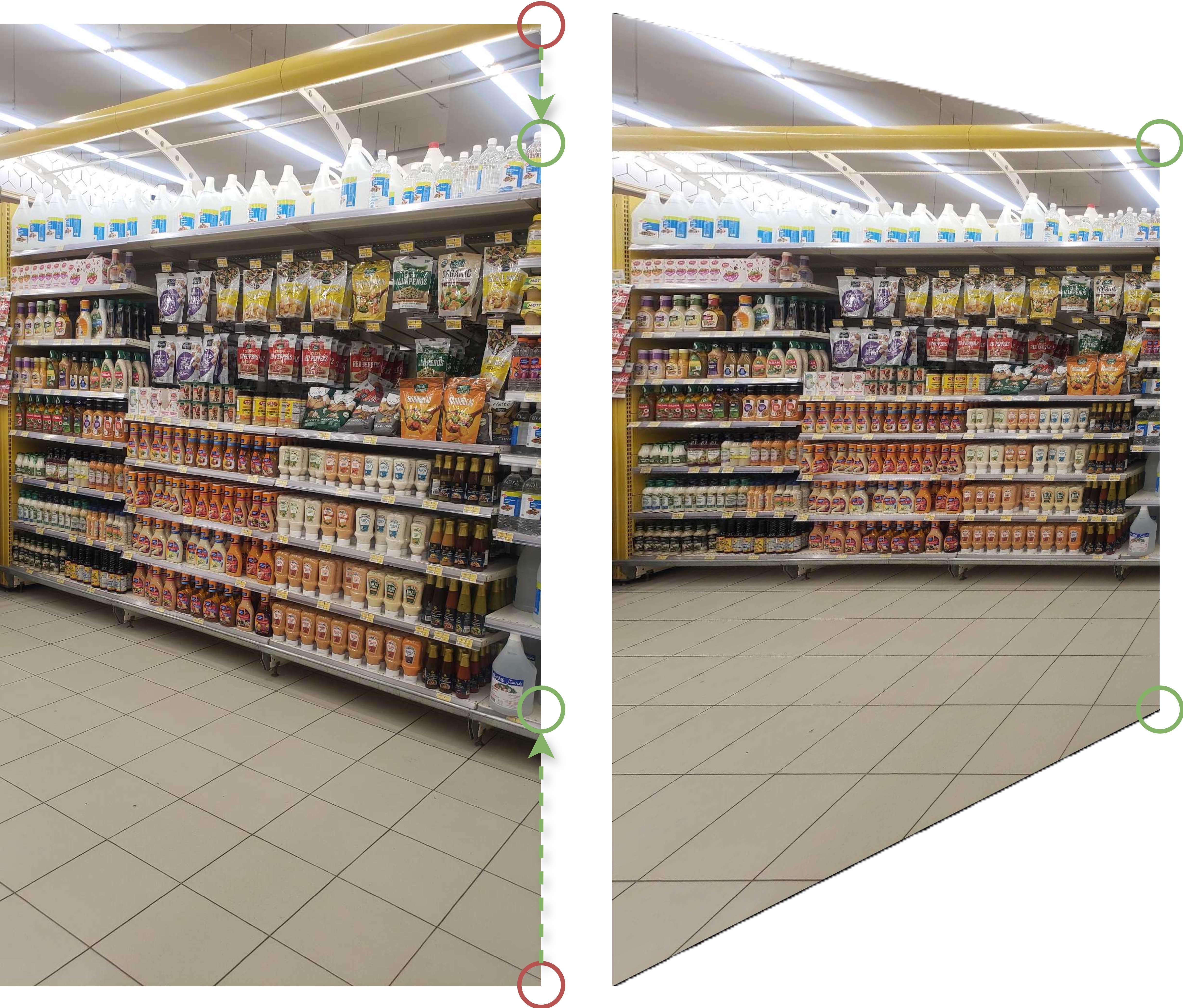}}
   \caption{The left image displays the original single-viewpoint capture with initial corner placements marked by red circles. On the right, the green circles indicate the interactively adjusted corner locations used to compute the rectifying homography, transforming the image to a fronto-parallel perspective.}
   \label{fig:dataset_generation}
\label{fig}
\end{figure}

To establish the ground truth for this task, we use a manual annotation process. Different from the 8-point (x,y) parameterization used by \cite{detone2016deep}, our annotation focuses only on the vertical displacement of the four corner points. This is because our primary objective is to make the horizontal shelf lines parallel. To simulate a realistic single-viewpoint perspective, annotators are constrained to adjust the vertical positions of the corners on only one side of the image at a time (either the left or the right), while the corners on the opposite side remain fixed. This approach defines the ground truth by mapping the two resulting vertical corner coordinates of the distorted shelf area to their rectified positions, as illustrated in Fig.~\ref{fig:dataset_generation}. The choice between left and right side displacement is uniformly distributed, ensuring the ShelfRectSet is balanced in this regard. As depicted in Fig.~\ref{fig:mean_displacement}, the average corner displacement across the ShelfRectSet is 19.16 pixels. To the best of our knowledge, this is the first dataset created for converting single-view images to their fronto-parallel versions. The complete ShelfRectSet dataset will be made publicly available.

As detailed in Table~\ref{tab:dataset_summary}, all images are resized to a uniform 224x224 resolution before being processed by the network. Since our ground-truth data is annotated on the original, variable-sized images, we adjust the corresponding homography for the resized input. This is achieved by converting the 4-point parameterization to its equivalent 3x3 homography matrix, $H_{orig}$, and then applying a scaling transformation. The new homography for the resized image, $H_{new}$, is calculated as:

\begin{equation}
H_{new} = S \cdot H_{orig} \cdot S^{-1}   
\end{equation}

where $S$ is the scaling matrix that transforms coordinates from the original image dimensions to the new dimensions, and $S^{-1}$ is its inverse. S defined as:

\begin{equation}
S = \begin{pmatrix} 
\frac{W_{new}}{W_{orig}} & 0 & 0 \\
0 & \frac{H_{new}}{H_{orig}} & 0 \\
0 & 0 & 1 
\end{pmatrix}
\end{equation}

\begin{table}[htbp]
\caption{ShelfRectSet statistics and configuration.}
\begin{center}
\begin{tabular}{|l|r|}
\hline
\multicolumn{2}{|c|}{\textbf{Dataset Details}} \\
\cline{1-2} 
\textbf{\textit{Parameter}} & \textbf{\textit{Value}} \\
\hline
Image Size (pixels) & 224 x 224 \\ \hline
Training Images & 8025 \\
Validation Images & 2259 \\
Test Images & 1139 \\ \hline
\textbf{Total Images} & \textbf{11423} \\
\hline
\multicolumn{2}{|c|}{\textbf{Displacement Balance}} \\
\cline{1-2} 
Left Corner Displacements & 5944 \\ \hline
Right Corner Displacements & 5479 \\
\hline
\end{tabular}
\label{tab:dataset_summary}
\end{center}
\end{table}

A principal challenge in training deep networks for this task is the scarcity of annotated data. Manually annotating a sufficiently large and diverse dataset to ensure model generalization is prohibitively expensive and labor-intensive. To overcome this, we introduce a novel data augmentation strategy designed to synthetically yet realistically expand the training set. This method avoids the need for paired multi-view supervision and allows us to generate a broad spectrum of geometric distortions from a limited set of seed images.

Our augmentation procedure is applied on-the-fly during training. With a given probability, an image selected from the training batch is first unwarped to its ground-truth fronto-parallel state. Then, a random 4-point homography matrix is sampled from our training dataset's distribution and applied to this unwarped image, creating a new, synthetic training sample with a known ground-truth label. This process exposes the model to a continuous and diverse stream of geometric distortions, which empirically reduces overfitting and enhances the model's ability to generalize to new, unseen images.

\section{Method}
\label{sec:model_architecture}

For feature extraction, our model employs a ConvNeXt-Nano \cite{liu2022convnet} backbone. We initialize the network with weights pretrained on ImageNet-12k and subsequently fine-tuned on ImageNet-1k. This architecture was chosen for its demonstrated strength in capturing rich, hierarchical feature representations. 

Our network's design is tailored to the specific problem of correcting vertical shelf distortion. While based on the 4-point parameterization from \cite{detone2016deep}, we simplify the regression task. Instead of regressing eight scalar values for the full (x,y) coordinates of the four corners, our model predicts only the four vertical (y) displacements. This strategic reduction focuses the network's learning capacity entirely on correcting the primary vertical distortion, which is essential for making the horizontal shelf lines parallel.

\begin{equation}
\mathcal{L}_{L2} = \frac{1}{N} \sum_{i=1}^{N} (y_i - \hat{y}_i)^2
\label{eq_l2}
\end{equation}

\begin{equation}
\mathcal{L}_{L1} = \frac{1}{N} \sum_{p} | I_1(p) - I_2(p) |
\label{eq_photometric}
\end{equation}

To further improve performance, we adopt two key strategies. First, to enhance training stability, we use normalized coordinate regression. All predicted corner coordinates are normalized to a [-1, 1] range before being passed to the loss function. This technique standardizes the regression target space, which is known to promote faster convergence and more reliable predictions. Second, the network is supervised by a composite loss function, defined as the weighted sum of two components: an L2 \eqref{eq_l2} loss on the predicted corner coordinates and a photometric reconstruction loss \eqref{eq_photometric}. This encourages the model to learn transformations that are not only geometrically accurate but also preserve the visual content of the shelf.

\section{Evaluation}
\label{sec:evaluation}

\begin{table*}[htbp]
\caption{Performance comparison against classical and deep learning approaches. Our model, ShelfRectNet, demonstrates superior accuracy.}
\label{tab:my_results}
\begin{center}
\begin{tabular}{|l|c|c|}
\hline
\multicolumn{3}{|c|}{\textbf{Performance Comparison}} \\
\cline{1-3} 
\textbf{\textit{Method}} & \textbf{\textit{Mean Corner Error (pixels)}} & \textbf{\textit{Inference Speed (ms)}} \\
\hline
Chaudhury et al. \cite{chaudhury2013autorectification} & 1.49 & 1279 \\ \hline
Deep Image Homography Net \cite{detone2016deep} & 3.17 & 2.311 \\ \hline
Deep Image Homography Net with ConvNext  & 1.397 & 2.172 \\ \hline

\textbf{ShelfRectNet (Ours)} & \textbf{\bestTestScore{}}& \textbf{2.172} \\
\hline
\end{tabular}
\end{center}
\end{table*}

We evaluated our framework on a held-out test set of real-world retail shelf images. Performance is measured by the mean corner error, defined as the average L2 distance in pixels between the predicted and ground-truth corners.

We benchmarked our model against a classic computer vision method \cite{chaudhury2013autorectification} and the foundational deep learning approach of \cite{detone2016deep}. Other learning-based methods were considered but ultimately excluded for several reasons. For instance, we did not adapt the method from Kang et al. \cite{Kang2019Combining} for a single-image view because its proposed network is less powerful than the VGG-style HomographyNet , and it is not a fully end-to-end approach as photometric loss is used for refinement rather than for updating the model. Other deep learning models were also unsuitable as they feature special modules and designs for multi-view or image-pair inputs.

To ensure a fair and insightful comparison, we made the following adjustments:

\textbf{Traditional Method}: Classical computer vision methods can fail under challenging real-world conditions like poor lighting or reflective surfaces, leading to large prediction errors. Due to high error rates on downscaled images, the method from Chaudhury et al. \cite{chaudhury2013autorectification} was tested using the original image sizes. To prevent severe failures from skewing the average, any prediction with an MCE above 45 pixels was excluded from the final calculation.

\textbf{Deep Learning Method}: To isolate the benefits of our design choices (e.g., augmentation, normalization) from the power of the backbone, we implemented two versions of the DHN network, adapted for single-view processing. The first is a direct adaptation, while the second replaces the original VGG-style backbone with our same ConvNeXt architecture. Unless otherwise stated, all deep learning baselines utilize the same feature backbone for a direct comparison.

\subsection{Implementation Detail}
All deep learning models were trained for 51 epochs using a batch size of 80 on RGB images resized to 224x224 pixels. Our data augmentation was applied with a probability of 0.5 to each training sample. We used the AdamW optimizer with a learning rate of 0.0001, a weight decay of 0.0001, and betas of (0.9, 0.999). A cosine annealing scheduler adjusted the learning rate down to a minimum of 1e-6, and the weight for the photometric loss component was set to 1.0. All evaluations were performed on a single Nvidia GeForce RTX 3070 GPU.

\subsection{Results}
\begin{figure*}[htbp]
    \centering 

    \begin{subfigure}{0.49\textwidth}
        \centering
        \includegraphics[width=\linewidth, height=0.40\textheight, keepaspectratio]{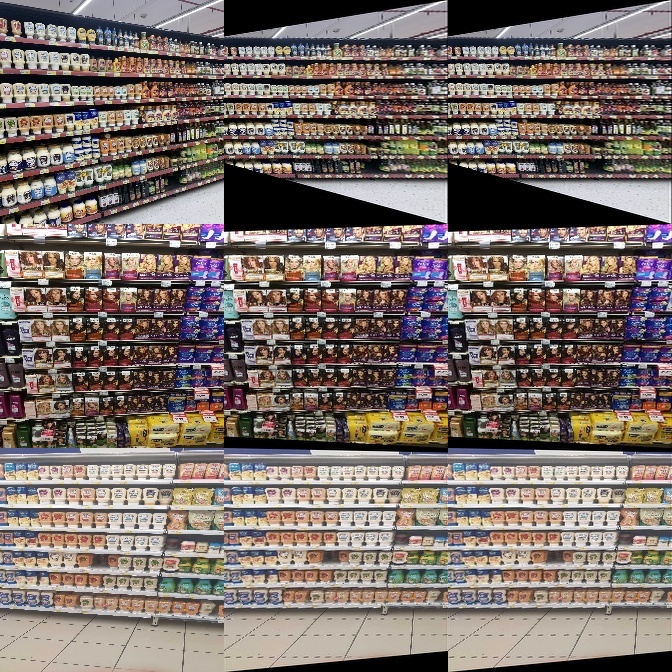}
        \subcaption{Examples of the model's best-performing results (lowest mean corner error).}
        \label{fig:best_images}
    \end{subfigure}
    \hspace{\fill} 
    \begin{subfigure}{0.49\textwidth}
        \centering
        \includegraphics[width=\linewidth, height=0.40\textheight, keepaspectratio]{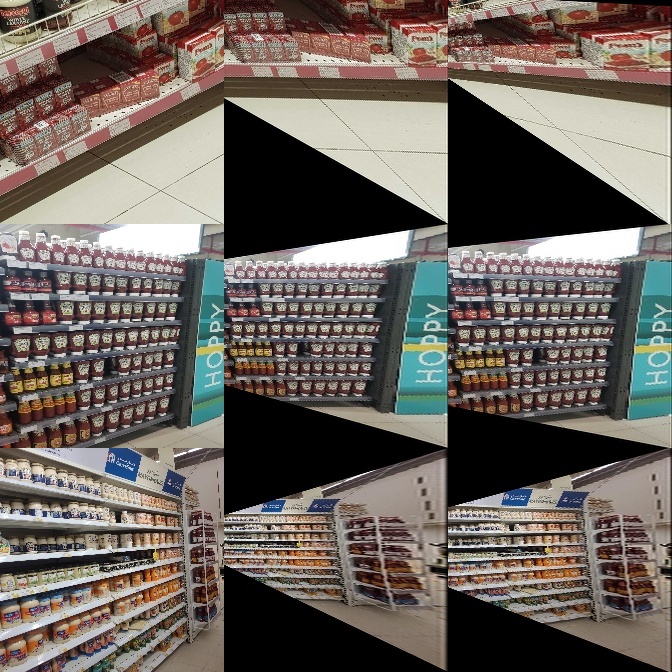}
        \subcaption{Examples of the model's most challenging cases (highest mean corner error).}
        \label{fig:worst_images}
    \end{subfigure}

    \caption{Qualitative comparison of our model's predictions against the ground truth. Each group from left to right displays the input image, the ground-truth homography overlay, and our model's predicted homography overlay.}
    \label{fig:horizontal_figure_group}
\end{figure*}

As shown in Table~\ref{tab:my_results}, our model, ShelfRectNet, achieves a state-of-the-art mean corner error of \bestTestScore{} pixels, a new  strong baseline on the benchmark ShelfRectSet, demonstrating competitive performance in both accuracy and inference speed. To provide a qualitative view of this performance, Fig.~\ref{fig:horizontal_figure_group} visualizes several predictions, showcasing examples of our model's best and worst-performing cases.

Our experiments reveal that the performance gain comes not just from using a powerful ConvNeXt backbone, but also from our specific methodological choices. By comparing the results of the two adapted DHN models, we can see that our full ShelfRectNet model outperforms both. This indicates that our design choices regarding range normalization, data augmentation, output parameterization, and the composite loss function all contribute significantly to the superior outcome, as further validated in the following ablation studies.

\section{Ablation Studies}
To validate our design choices, we conducted a series of ablation studies analyzing the impact of key components of our framework.

\begin{table}[htbp]
    \centering
    \caption{Ablation studies on our model's components. The first row shows the performance of our full model, and subsequent rows show the performance when a specific component is removed or altered.}
    \label{tab:ablation_single_col}
    \begin{tabular}{|l|c|}
        \hline
        \textbf{Configuration} & \textbf{MCE (pixels)} \\
        \hline\hline
        \textbf{Our Full Model (Baseline)} & \textbf{\bestTestScore{}} \\
        \hline
        Without Photometric Loss & 1.445 \\
        Without Augmentation & 1.338 \\
        Without Range Normalization & 1.430 \\
        Using Grayscale Input & 1.361 \\
        \hline
    \end{tabular}
\end{table}

\begin{table}[htbp]
\caption{Ablation study comparing the 4-point and 3-point network parameterizations.}
\label{tab:ablation_results}
\begin{center}
\begin{tabular}{|l|c|c|}
\hline
\multicolumn{3}{|c|}{\textbf{Ablation Study: Output Parameterization}} \\
\cline{1-3} 
\textbf{\textit{Method}} & \textbf{\textit{Validation MCE }} & \textbf{\textit{Test MCE }} \\
\hline
\textbf{4-Point Network (Ours)} & 1.284 & \textbf{\bestTestScore{}} \\ \hline
3-Point Network&  \textbf{1.235} & 1.301 \\
\hline
\end{tabular}
\end{center}
\end{table}

\subsection{Photometric Loss}

To assess the contribution of the photometric loss, we trained an identical model with this component disabled, relying solely on the L2 corner loss.

The results, summarized in Table~\ref{tab:ablation_single_col}, show that removing the photometric loss led to a noticeable degradation in performance. The model trained without it achieved a higher corner error on the test set. 

\subsection{Alternative Network Output Parameterization}

In our data annotation process, only the vertical coordinates of the corners on one side of the image (either left or right) are adjusted, while the other side remains fixed. This implies that in our 4-point parameterization $(y_1, y_2, y_3, y_4)$, two of the outputs are always zero. This inherent redundancy suggested an alternative network design could be more efficient.

We explored a model that predicts a 3-point output instead of a 4-point one. This network was designed with a multi-task head: one binary classification output to predict which side's corners (left or right) were displaced, and two regression outputs for the corresponding non-zero vertical displacement values.

Interestingly, this 3-point network demonstrated superior performance during training, achieving lower loss and higher accuracy on the validation set compared to our original 4-point network. However, this improvement did not generalize to the unseen test data. As shown in Table~\ref{tab:ablation_results}, the final mean corner error on the test set was higher for the 3-point network, indicating that the original 4-point parameterization, despite its redundancy, provides better regularization and leads to a more robust model.

\subsection{Model Size and Pretraining}

To assess the impact of model scale and the benefits of transfer learning, we conducted an ablation study comparing different model configurations. We evaluated four primary variants: our main approach using a ConvNeXt-Nano backbone with ImageNet-12k pretraining and ImageNet-1k fine-tuned, the same ConvNeXt-Nano model trained from scratch without any pretraining, a larger ConvNeXt-Tiny model and ConvNext-Pico model trained with ImageNet-1k. The Tiny model followed the same pretraining strategy as our primary model, being pretrained on ImageNet-12k and subsequently fine-tuned on ImageNet-1k. For the larger Tiny model, the learning rate was multiplied by 40 to ensure stable training and batch size decreased to 40 to fit the model to memory. The results of these experiments are summarized in Table~\ref{tab:ablation_model_size}.

As shown in the table, pretraining provides a significant advantage. The larger ConvNeXt-Tiny model, with 27.8M parameters, shows a slight performance advantage over the smaller Nano variant. However, considering the substantial increase in model size, our primary approach using the pretrained ConvNeXt-Nano model offers the best trade-off between accuracy and computational cost. Due to GPU memory limitations, we did not extend our experiments to models larger than the ConvNeXt-Tiny.

\begin{table}[htbp]
\caption{Ablation on model size and pretraining.}
\label{tab:ablation_model_size}
\begin{center}
\begin{tabular}{|l|c|c|}
\hline
\multicolumn{3}{|c|}{\textbf{Ablation: Model Size \& Pretraining}} \\
\cline{1-3} 
\textbf{\textit{Model}} & \textbf{\textit{Params (M)}} & \textbf{\textit{Test MCE}} \\
\hline

ConvNeXt-Pico (IN-1k) & 8.5 & 1.43 \\ \hline
ConvNeXt-Nano (Random Initialization) & 15.0 & 2.475 \\ \hline
ConvNeXt-Nano (IN-12k) & 15.0 & 1.350 \\ \hline
\textbf{ShelfRectNet (Ours)} & \textbf{15.0} & \bestTestScore{} \\ \hline
ConvNeXt-Tiny (IN-12k + IN-1k) & 27.8 & \textbf{1.292} \\

\hline
\end{tabular}
\end{center}
\end{table}

\subsection{Augmentation}

To quantify impact of augmentation strategy, we trained a baseline model without this augmentation procedure. The results, detailed in Table~\ref{tab:ablation_single_col}, clearly show that our augmentation strategy is crucial for generalization. 

\subsection{Range Normalization}

To improve training stability and predictive consistency, we normalize the target corner coordinates to a [-1, 1] range before they are passed to the loss function. We conducted an ablation study to verify the effectiveness of this strategy by training an identical model without this normalization step. The results, presented in Table~\ref{tab:ablation_single_col}, confirm the benefits of this approach. The model trained with normalized coordinates achieved a lower mean corner error.

\subsection{Grayscale}

To determine the importance of color information for our task, we conducted an ablation study comparing the performance of our model trained on standard RGB images versus grayscale images. As shown in Table~\ref{tab:ablation_single_col}, the model trained on RGB images achieved a lower mean corner error. 

\section{Future Work}

One promising direction is to explore more advanced data augmentation strategies to further enhance model robustness and generalization. While our current method of sampling homographies from the training distribution has proven effective, we plan to investigate adaptive augmentation techniques such as RandAugment \cite{RandAug}. This would involve creating a pool of diverse geometric and photometric transformations and learning an optimal sampling policy instead of applying them with fixed probabilities. By modeling the distribution of augmentation strengths, potentially using a Gaussian distribution for the 4-point parameterization, the model could be trained on a more targeted and effective curriculum of synthetic data. This approach would allow us to systematically explore a wider range of challenging yet realistic scenarios, potentially leading to even lower corner errors and improved performance in real-world deployments.

\section{Conclusion}

In this paper, we introduced ShelfRectNet, a deep learning framework designed for the practical challenge of rectifying retail shelf images from a single, arbitrary viewpoint. By leveraging a ConvNeXt backbone, a targeted 4-point homography parameterization, and a novel data augmentation strategy that simulates realistic perspective distortions, our model achieves state-of-the-art performance. With a mean corner error of \bestTestScore{} pixels on our new benchmark dataset, ShelfRectSet, our method proves to be both accurate and efficient. By successfully addressing the single-view rectification problem and making our dataset and code publicly available, we hope to facilitate further research and development in automated retail monitoring and other domain-specific geometric vision tasks.

\bibliographystyle{IEEEtran}
\bibliography{references} 

\end{document}